\documentclass{article}

\usepackage[numbers]{natbib}
\usepackage[preprint]{neurips_2026}

\usepackage[utf8]{inputenc}
\usepackage[T1]{fontenc}
\usepackage{hyperref}
\usepackage{url}
\usepackage{booktabs}
\usepackage{amsfonts}
\usepackage{amsmath}
\usepackage{amssymb}
\usepackage{amsthm}
\newtheorem{theorem}{Theorem}
\newtheorem{proposition}[theorem]{Proposition}

\newtheorem{assumption}{Assumption}
\newtheorem{definition}{Definition}

\usepackage{nicefrac}
\usepackage{microtype}
\usepackage[table]{xcolor}
\usepackage{wrapfig}
\usepackage{graphicx}
\usepackage{algorithm}
\usepackage{algpseudocode}
\usepackage{CJK}
\usepackage{multirow}
\usepackage{array}
\usepackage{enumitem}
\usepackage{listings}
\usepackage{tikz}
\usetikzlibrary{positioning, arrows.meta, shapes.geometric, calc, backgrounds}
\lstdefinestyle{regex}{
  basicstyle=\scriptsize\ttfamily,
  breaklines=true,
  breakatwhitespace=false,
  columns=fullflexible,
  frame=none,
  showstringspaces=false,
  keepspaces=true,
  upquote=true,
  aboveskip=2pt,
  belowskip=2pt,
  xleftmargin=2em
}
\usepackage{subcaption} 

\let\oldparagraph\paragraph
\renewcommand{\paragraph}[1]{\oldparagraph{#1}}

\title{Interpretable Discriminative Text Representations via Agreement and Label Disentanglement}

\author{ \href{https://orcid.org/0000-0001-8687-4208}{\includegraphics[scale=0.06]{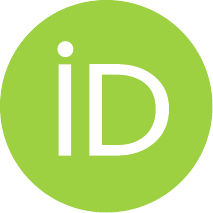}\hspace{1mm}Tong Wang} \\ 
	School of Management\\
	Yale University\\
	New Haven, CT, US, 06511 \\
	\texttt{tong.wang.tw687@yale.edu} \\
	\And
	\href{https://orcid.org/0000-0003-2041-6671}{\includegraphics[scale=0.06]{orcid.pdf}\hspace{1mm}Yiqing Xu} \\
	Department of Political Science\\
	Stanford University\\
	Stanford, CA, US, 92093 \\
	\texttt{yiqingxu@stanford.edu} \\
    \And
	\href{https://orcid.org/0000-0002-1393-5417}{\includegraphics[scale=0.06]{orcid.pdf}\hspace{1mm}Leo Yang} \\
	Department of Accountancy, Economics and Finance\\
	Hong Kong Baptist University\\
	Kowloon, Hong Kong SAR \\
	\texttt{leoyang@hkbu.edu.hk} \\
}

\begin{document}
\begin{CJK}{UTF8}{gbsn}

\maketitle

\begin{abstract}
Interpretable text representations should expose coordinates that are not only predictive, but also meaningful enough for independent auditors to apply. Existing discriminative representations often use anonymous embedding directions, while concept-bottleneck and LLM-assisted methods attach natural-language names to features without ensuring that those definitions are reproducible or distinct from the target label. We propose an operational criterion for interpretable discriminative text representations: each coordinate should satisfy conceptual clarity, measured by chance-adjusted agreement between independent annotators applying the feature definition, and label disentanglement, meaning the feature should not merely paraphrase the prediction target. We instantiate this criterion in LLM-assisted Feature Discovery (LFD), an iterative method that proposes lexical and semantic features from contrastive outcome-opposed text pairs, screens candidates using cross-LLM Cohen's $\kappa$, and selects features by residual held-out predictive gain. A stylized analysis connects the $\kappa$ screen to a per-feature annotation-noise bound, formalizing agreement as a reliability check. Across ten text-classification tasks spanning seven corpora, LFD matches the predictive performance of a strong text bottleneck baseline while producing substantially clearer and less label-entangled features. Human audits with 232 raters show that LFD features achieve higher human--human and human--LLM agreement than baseline concepts, and raters consistently judge them as less label-leaking. These results suggest that agreement-tested, label-disentangled coordinates provide a practical auditability standard for interpretable text classification.\end{abstract}

\clearpage
\section{Introduction}
\label{sec:introduction}

What does it mean for a learned representation of text to be
\emph{interpretable}? Existing coordinate-level methods often make
one of two compromises. Anonymous discriminative bases ---
principal component analysis (PCA), sparse probes, and
task-aligned embeddings --- can predict
well, but their coordinates are not named constructs an analyst can
read or apply. Conversely, concept-bottleneck and large-language-model
(LLM)-assisted pipelines attach natural-language names to features,
but often
validate interpretability only post hoc: the resulting definitions
may sound plausible while remaining too abstract, ambiguous, or
label-entangled to support auditing or intervention
\citep{koh2020concept, ludan2024interpretable, belinkov2022probing}.
A feature name is not enough. To audit a model's reliance on a
coordinate, or to intervene on that coordinate, an analyst must know
what the feature reliably picks out in new texts.

Social science offers a stricter standard. Researchers coding
documents for latent constructs (e.g.\ \emph{aggressive language},
\emph{content deemed sensitive by authorities}, \emph{deceptive review}) write an operational
codebook, ask independent coders to apply it, and report
chance-adjusted inter-rater agreement (Cohen's
$\kappa$~\citep{cohen1960coefficient} or Krippendorff's
$\alpha$~\citep{krippendorff2018content}) as evidence that the
construct is reproducibly measurable; substantial
agreement~\citep{landis1977measurement}, as known as ``construct validity'' in social science~\citep{cronbach1955construct}, is a precondition for
scientific use. We import this standard into representation
learning: a coordinate is conceptually clear only if its
natural-language definition transfers across independent annotators
with substantial chance-adjusted agreement.

We call this requirement \textbf{conceptual clarity}, but it is not
sufficient. A feature whose definition simply restates the target
--- e.g.\ ``mention of sports terminology'' for an is-this-about-sports
classifier --- is easy to apply but explains nothing; the bottleneck
collapses to a renamed classifier. We therefore require a second
property: \textbf{label disentanglement}, meaning the feature's
information content is distinct from the target itself.
Interpretability admits other readings (faithfulness, mechanistic
causality, sparsity, $\ldots$); we propose one falsifiable kernel
on top of whatever else a representation is asked to satisfy ---
each coordinate must be both conceptually clear and
label-disentangled --- as an auditability standard, not the final
definition.

We instantiate the two-property definition in
\textbf{LFD} (LLM-assisted Feature Discovery). At each round, a
\emph{proposer} LLM is shown a small minibatch of contrastive,
outcome-opposed text pairs and asked to abduce (in
Peirce's sense of inference to the best
explanation~\citep{burks1946peirce}) named predicates that
explain the contrast. Proposals come through two
lanes: a \emph{lexical} lane producing surface-form features
(e.g., \textit{``contains a final game score''}), and a
\emph{semantic} lane producing reading-comprehension rules
(e.g., \textit{``invokes a statute without giving a personal narrative''}).A labeler
applies the feature to annotate the instances, and then a separate, independent
\emph{examiner} LLM applies the same feature definition and codebook to annotate the instances independently. Only
candidates whose label vectors from two independent LLMs achieve a large enough cross-LLM
$\kappa$ are kept for further selection.  Conceptual clarity is enforced by this
cross-LLM agreement screen: a candidate definition that does not
transfer to an independent examiner is rejected before predictive
gain is considered, ruling out the single-LLM definer-applier
closed loop of \citet{gilardi2023chatgpt, tornberg2023chatgpt4}
in which one LLM both defines and applies a feature. The same screen
also has a statistical consequence:
Proposition~\ref{thm:kappa_generalization} (a stylized, coordinate-wise
result) ties cross-rater $\kappa$ to a per-feature noise-rate bound,
formalising the intuition that unreproducible features are
unreliable supervision. Label disentanglement is encouraged structurally:
contrastive minibatches ask what distinguishes outcome-opposed texts,
pushing proposals toward local cues rather than label paraphrases,
and residual-gain selection admits a feature only if it explains
held-out signal not already captured by previous features. 

\paragraph{Contribution.}
The main contribution of this paper is an operational definition of
interpretability for learned text representations. We argue that an
interpretable coordinate should satisfy two testable properties: conceptual clarity and label disentanglement. LFD is an
instantiation of this definition: it uses cross-LLM agreement to
screen for clarity and contrastive residual selection to discourage
label paraphrases. Our theoretical result is a stylized noise-rate bound that
justifies the $\kappa$-screen as a per-feature reliability check
rather than a multivariate generalization guarantee. Empirically,
we run extensive experiments across ten text-classification tasks
spanning seven corpora 
and complement the automated
evaluation with a human audit in which independent raters
score every LFD and Text Bottleneck Model (TBM) feature for definitional clarity and
label-disentanglement, validating that the proposed properties hold in
practice.

\section{Related Work}
\label{sec:related}

For text classification, two desiderata pull on a learned
representation: coordinates should carry the discriminative
signal of the outcome (\emph{useful}) and each should carry a
definition an analyst can read and apply (\emph{auditable}).
Existing methods occupy a 2$\times$2 grid
(Table~\ref{tab:2x2}); most achieve at most one.

\begin{table}[h]
\centering
\small

\setlength{\tabcolsep}{4pt}
\renewcommand{\arraystretch}{1.1}
\caption{The 2$\times$2 organization of text representation
methods.}
\label{tab:2x2}
\begin{tabular}{l|p{5.5cm}|p{5.2cm}|}
\multicolumn{1}{c}{} & \multicolumn{1}{c}{\textbf{Anonymous coordinates}} & \multicolumn{1}{c}{\textbf{Named (interpretable)}} \\
\cline{2-3}
\textbf{Discriminative} & PCA-on-$\boldsymbol{\Sigma}_y$, sparse probe, fine-tuned BERT & \textbf{LFD (this paper)}, TBM, CBM, LaBo \\
\cline{2-3}
\textbf{Non-discriminative} & full embedding, vanilla PCA & LDA, BERTopic, keyword dictionaries \\
\cline{2-3}
\end{tabular}
\end{table}

\paragraph{Anonymous or non-discriminative coordinates.}
PCA on outcome-weighted moments, linear and sparse probes
\citep{belinkov2022probing, dalvi2020analyzing}, fine-tuned
encoders, and task-vector arithmetic
\citep{ilharco2022editing, park2023linear, aghajanyan2020intrinsic,
mu2018allbutthetop} recover task-aligned directions in pretrained
representations. The coordinates are linear combinations of
embedding dimensions and carry no human-readable label: strong
predictors, weak audit surfaces. They are discriminative but anonymous. On the other hand, bag-of-words with sparse linear models
\citep{tibshirani1996regression}, keyword dictionaries
\citep{loughran2011liability, pennebaker2001linguistic}, and topic
models \citep{blei2003latent, roberts2014structural,
grootendorst2022bertopic} yield coordinates with human-readable
descriptions, but each is unsupervised at the
coordinate-construction stage --- the features index language
structure rather than outcome variation, and align with outcomes
only incidentally.

\paragraph{Interpretable discriminative coordinates: concept-bottleneck
models.} Concept-bottleneck models (CBM)~\citep{koh2020concept}
predict through a layer of named concepts. Recent work generates
the concept set with an LLM: in vision,
LaBo~\citep{yang2023language}, label-free
CBMs~\citep{oikarinen2023label}, and post-hoc
CBMs~\citep{yuksekgonul2023post}; the closest text-side analogue
is TBM~\citep{ludan2024interpretable}, where a single LLM
iteratively proposes concepts on misclassified examples, measures
them, and a linear head predicts. We differ structurally:
\emph{proposer and measurer must be distinct LLMs, and a candidate
is retained only if cross-LLM $\kappa \geq 0.70$}. The CBM
literature flags information leakage and unfaithful concept
measurement as central
threats~\citep{mahinpei2021promises, ramaswamy2022overlooked};
the single-model closed loop is their structural source. Our
screen breaks that specific loop, with two consequences: (i)
the head can be any $\boldsymbol{\phi}$-measurable function, since
the audit surface is the concept layer; (ii) interpretability
becomes testable per-feature during training, not post-hoc.
A second structural distinction is \emph{when} the LLM is queried.
TBM, LaBo, and label-free CBMs operate in an
\emph{LLM-as-classifier} regime: each concept value at inference
requires re-querying the LLM, so the bottleneck contains $K$
sub-classifications, each as opaque as the original label. LFD's
two-lane design (regex deterministic + cross-LLM-validated
labeling spec) shifts to an \emph{LLM-as-interpreter} regime: the
LLM contributes at design time, and inference is a pure function
of the text. The cross-LLM $\kappa$-screen is the operational
test that distinguishes the two regimes --- a feature whose value
is the proposer's holistic judgment cannot be reproduced by an
independent rater at high $\kappa$ without re-running the same
LLM.

\section{Problem Setup and Method}
\label{sec:method}

\subsection{Interpretable Discriminative Representations}

Let $\{(x_i, y_i)\}_{i=1}^N$ be labeled text with $y_i \in \{0,1\}$.
A \emph{named feature} is a pair $f = (\phi_f, \rho_f)$ where
$\phi_f: \mathcal{X}\to\{0,1\}$ is a labeling function and $\rho_f$
is a natural-language operational definition sufficient for an
independent annotator to apply $\phi_f$ consistently. Each accepted
feature carries a structured codebook record.

\begin{definition}[An operational interpretability criterion]
\label{def:interpretable}
Interpretability admits many readings (faithfulness, mechanistic
causality, sparsity, simulatability, $\ldots$). We propose
\emph{one} falsifiable, pre-hoc standard that any interpretable
text-classification feature should arguably meet, leaving the
broader question open. Given thresholds $\kappa^* \in [0,1]$ and
$\tau \in [0,1]$, we say a named feature $f = (\phi_f, \rho_f)$
satisfies our \emph{operational interpretability criterion} iff:
\begin{enumerate}[leftmargin=*,label=(\roman*)]
  \item \textbf{Conceptual clarity.}
    $\kappa(f) \geq \kappa^*$, where $\kappa(f)$ is Cohen's $\kappa$
    between two independent annotators given only $\rho_f$.
  \item \textbf{Label disentanglement.}
    $|\rho(f, y)| \leq \tau$ on a held-out sample, where
    $\rho(f, y)$ is the centered cosine (Pearson correlation) between
    the feature's labeling vector and the label vector.
\end{enumerate}
\S\ref{sec:disentanglement} reports two complementary diagnostics
beyond (ii) --- definition-level cosine to the label question and a
human disentanglement rubric --- which probe label-entanglement at
the definition level rather than the labeling-vector level.
\end{definition}

\begin{definition}[Interpretable discriminative representation]
\label{def:idr}
A set $\mathcal{F} = \{f_1, \ldots, f_K\}$ of named features is an
\emph{interpretable discriminative representation} iff every $f_k$
is interpretable in the sense of
Definition~\ref{def:interpretable}, and $\mathcal{F}$ is
\textbf{discriminative}: its selection is supervised by $\{y_i\}$
and each $f_k$ contributes positively to held-out predictive
accuracy of $y$ given the rest of the basis.
\end{definition}

In this paper we choose $\kappa^* = 0.70$ to match the
substantial-agreement bar of \citet{landis1977measurement}. We
\emph{intentionally} set $\tau = 1$ --- i.e., disable the
disentanglement gate --- to test whether LFD's structural design
alone (matched contrastive minibatches, contrastive abductive
reasoning, residual-gain selection) produces label-disentangled
features without a post-hoc filter. Empirically the answer is yes:
max realized $|\rho| = 0.59$ across all admitted features
(\S\ref{sec:disentanglement}, Loan Application's
standard-template regex), naturally below the
substantial-agreement bar $\tau = 0.60$. In practice $\tau$ acts
as a per-feature \textbf{granularity dial}: a tighter $\tau$
pushes the basis toward narrow, decompositional sub-features that
combine multiplicatively; a looser $\tau$ admits broader,
label-correlated concepts (in the limit, TBM-style coordinates
that approximate the label directly).

\paragraph{Auditability.} A representation satisfying
Definition~\ref{def:idr} is auditable in the operational sense an
analyst or regulator needs: each coordinate carries a fixed
natural-language rule that any subsequent rater can apply to new
documents, without access to model state, training code, or pipeline
internals. This is the precondition for downstream review,
contestability, and transfer.

\subsection{LFD}
\label{sec:method_pipeline}
\label{sec:proposal}
\label{sec:screen}

LFD discovers an interpretable discriminative representation
iteratively. Each round draws a contrastive minibatch of
outcome-opposed text pairs, prompts a proposer LLM to abduce a
named candidate that distinguishes them, has an independent
examiner LLM relabel the dataset to compute Cohen's $\kappa$ on
the candidate, and admits the candidate if it passes the
$\kappa$-gate and improves held-out predictive gain. 
The rest of this subsection walks
through each step in order, noting how each design choice
connects to the two interpretability properties of
Definition~\ref{def:idr}.

\paragraph{Pair sampling.} The pipeline runs in two phases. The
brief \emph{bootstrap} phase presents the proposer with two random
groups, one drawn from each label class, and asks for features
that distinguish Group A from Group B; the first 1--3 accepted
features typically capture the dominant predictive signal. Once
there's no improvement from the bootstrap phase, the iteration enters the \emph{residual}
phase, where each call shows a \emph{contrastive pair} of
outcome-opposed texts that are similarity-matched in a mid-range
band (TF-IDF for the lexical lane; embedding for the semantic
lane).\footnote{Top-tail similarity yields near-duplicates whose
contrast is a surface artifact; bottom-tail similarity shares too
little context.} The two texts are surface-similar but differ in
outcome --- analogous to a matched control--treatment pair where
holding context fixed isolates the effect of the remaining
variable. The proposer's contrastive question (an instantiation of
Mill's method of difference~\citep{mill1843system}) can only be
answered by the local cue that flipped the label, structurally
pushing proposals away from broad label-correlated patterns toward
fine-grained sub-cues. A diversity prompt complements the matching
by asking the proposer to target positive-class subgroups not yet
covered by admitted features. Together, matched-pair sampling and
the diversity prompt constitute the first structural mechanism
driving the basis toward label disentanglement.

\paragraph{Proposal.} The proposer LLM is invoked through two
lanes per call. The \emph{lexical lane} asks for surface-form
rules; each candidate carries a regex implementing the rule,
applied deterministically by any rater ($\kappa = 1$ by
construction). The \emph{semantic lane} asks for a step-wise
reading-comprehension procedure that an examiner can apply
mechanically. Both formats commit proposals to a rule that any
rater can run; holistic ``is the text $X$?'' framings whose
application depends on the rater's interpretive prior are ruled
out structurally. Without this constraint the $\kappa$-test below
would be ill-defined --- the two-lane requirement is what makes
conceptual clarity testable at all.

\paragraph{Examination.} A second, independent LLM (the
\emph{examiner}) receives each candidate definition $\rho_f$ and
relabels the dataset without access to either the proposer's
labels or the labeler's labels. We compute Cohen's $\kappa$
\citep{cohen1960coefficient} between the \emph{labeler} and
\emph{examiner} label vectors --- both LLMs that apply the same
definition to the same texts --- and discard any candidate with
$\kappa < \kappa^* = 0.70$, the substantial-agreement bar of
\citet{landis1977measurement}. This is the literal test of
conceptual clarity: a definition that does not transfer between
independent annotators fails. The screen also helps disentanglement
as a side effect --- abstract definitions that paraphrase the
label (``is the tone ironic?'', ``is this text aggressive?'') tend
to fail cross-LLM agreement on borderline cases, so what survives
is concrete enough to be applied mechanically and harder to make
extensionally equivalent to $y$. The cross-LLM split additionally
rules out the \emph{single-LLM definer-applier closed loop} of
LLM-as-annotator methods~\citep{gilardi2023chatgpt,
tornberg2023chatgpt4} and single-LLM concept-bottleneck pipelines,
where the same model defines and applies the feature; it does
\emph{not} rule out the broader failure mode of two LLMs sharing
residual blind spots from overlapping training data, which we
audit empirically in \S\ref{sec:clarity}. We
note one structural asymmetry between the lanes:
deterministic-regex features achieve $\kappa = 1$ by construction
(any two raters running the same regex produce identical labels),
so the cross-LLM screen does its empirical work on the
semantic-lane subset. 

\paragraph{Selection.} Survivors of the $\kappa$-gate are scored
on incremental held-out balanced accuracy under a downstream
classifier (LightGBM~\citep{ke2017lightgbm} by default). In the residual phase,
gain is computed against the residual of $y$ after prior accepted
features, not raw $y$-correlation. A candidate near-equivalent to
the label scores high on raw $y$-correlation but, once one such
candidate has been admitted, has near-zero residual gain;
subsequent restatements are rejected, and the basis grows into
orthogonal sub-features that each explain a local signal previous
features missed. Together with matched-pair sampling, this is the
second structural mechanism that drives the basis toward label
disentanglement. A candidate is admitted only if it additionally
satisfies the disentanglement gate $|\rho(f, y)| \leq \tau$ on the
held-out subset, ruling out near-paraphrase features whose firing
pattern alone is extensionally close to $y$ even when their
definition is clear.

The procedure
terminates when no candidate clears the gain threshold $\delta$;
a final pruning step removes any $f$ whose exclusion leaves
validation BA unchanged.

\subsection{Theoretical motivation: a stylized noise-rate bound for the $\kappa$-screen}
\label{sec:theory}

Beyond filtering unreproducible candidates, the $\kappa$-screen has
a statistical consequence: under a stylized noise model,
$\kappa \geq \kappa^*$ corresponds to a per-feature noise-rate cap
that, applied coordinate-wise, bounds the noise-induced inflation
of the downstream generalization gap. A candidate feature with
definition $\rho_f$ is associated with a \emph{latent target
labeling} $\phi^*_f: \mathcal{X}\to\{0,1\}$ --- the labels a
perfect rater applying $\rho_f$ would produce. Each LLM annotator
$A \in \{\mathcal{P}, \mathcal{E}, \mathcal{L}\}$ returns a noisy
version $A(x) = \phi^*_f(x) \oplus \epsilon^A(x)$, where
$\epsilon^A(x)$ is the rater's per-instance error.

\begin{assumption}[Annotation noise model]
\label{ass:annotation_noise}
For a fixed feature definition $\rho_f$ and two annotators
$A, B$:
\begin{enumerate}[leftmargin=*,label=(\roman*)]
\item (Symmetric and equal noise.)
  $\mathbb{P}(\epsilon^A(x) = 1 \mid \phi^*_f(x) = 0) =
   \mathbb{P}(\epsilon^A(x) = 1 \mid \phi^*_f(x) = 1) = \eta_A$,
  and $\eta_A = \eta_B = \eta$. The equal-rate condition is what
  lets cross-rater $\kappa$ bound the per-rater $\eta$; under unequal rates $\kappa$
  only constrains a joint reliability quantity.
\item (Conditional independence.)
  $\epsilon^A(x) \perp\!\!\!\perp \epsilon^B(x) \mid \phi^*_f(x)$.
\end{enumerate}
\end{assumption}

\begin{proposition}[Cross-rater $\kappa$ bounds per-feature annotation noise]
\label{thm:kappa_generalization}
Let $\mathcal{H}$ be a hypothesis class with VC dimension $d$ and
let $\mathbf{F} \in \{0,1\}^{N \times K}$ be the labeler's matrix
of $K$ LFD-admitted features. Suppose every $f_k$ passes the
cross-LLM screen at threshold $\kappa^*$, and
Assumption~\ref{ass:annotation_noise} holds. Then the empirical
risk minimizer $\hat h \in \mathcal{H}$ trained on $(\mathbf{F},
\mathbf{y})$ satisfies, with probability $\geq 1 - \delta$,
\[
R(\hat h) - R(h^\star) \;\leq\; \frac{1}{1 - 2\bar\eta}\cdot
\mathcal{O}\!\left(\sqrt{\frac{d + \log(1/\delta)}{N}}\right),
\qquad
\bar\eta \;\leq\; \frac{1 - \sqrt{\kappa^*}}{2},
\]
where $h^\star$ is the risk minimizer on the noise-free latent
matrix $\mathbf{F}^\star$, and $(1 - 2\bar\eta)^{-1}$ is the
noise-tolerance penalty of \citet{natarajan2013learning}. For
$\kappa^* = 0.70$, the inflation factor $\approx 1.20$.
The $\bar\eta$ bound holds uniformly in feature prevalence
$\pi = \mathbb{P}(\phi^*_f = 1)$.
The result is a single-feature reliability statement lifted to
the matrix setting by union bound over the $K$
coordinates~\citep{patrini2017makingdeep}; it is \emph{not} a
tight covariate-noise generalization bound for the joint
$K$-feature classifier, where dependencies between feature errors
and the head's effective complexity would require strictly more
careful treatment.
\end{proposition}

\paragraph{Scope.} The bound is stylized; its value is the chain
$\kappa \to \bar\eta \to (1-2\bar\eta)^{-1}$ and the operating
point $\kappa^*=0.70 \Rightarrow$ inflation $\leq 1.20$. Three
caveats. \textbf{(a)} Coordinate-wise + union bound, not a tight
multivariate covariate-noise bound. \textbf{(b)} Symmetric and
equal noise; under asymmetric noise the weighted Cohen's $\kappa$
replaces the symmetric form. \textbf{(c)} Conditional independence
is the most delicate assumption; we audit it empirically in
\S\ref{sec:clarity}, where $\kappa_{\text{ll}} - \kappa_{\text{hh}}
\approx +0.36$ on AG-Sci/Tech semantic features implies an
empirical inflation factor $\approx 1.30$--$1.45$ rather than the
idealized $1.20$.

\section{Experiments}
\label{sec:experiments}

The experiments answer three questions. First, how does LFD's
predictive performance compare across these representation families,
especially within the named/discriminative setting
(\S\ref{sec:main_results})? Second, are LFD features less
label-entangled than concepts produced by a single-LLM
concept-bottleneck pipeline (\S\ref{sec:disentanglement})? Third,
are LFD feature definitions conceptually clearer and more aligned with human understanding (\S\ref{sec:clarity})? 

\paragraph{Datasets.}
We evaluate on ten tasks across seven corpora: five
domain / social-science corpora (AdParaphrase
\citep{murakami2025adparaphrase}, Loan Application, CFPB Complaints, Dec.
Reviews~\citep{ott2011finding}, Gaslighting) and two standard NLP
benchmarks (iSarcasm~\citep{oprea2021isarcasm} and AG
News~\citep{zhang2015character} as four 1-vs-rest tasks). The Loan
Application corpus is proprietary; CFPB and Gaslighting are
crawled from public internet sources (CFPB from the U.S.\ Consumer
Financial Protection Bureau's public complaints database,
Gaslighting from a public social-media platform). All runs use
balanced 1{,}000-sample subsets.
The set deliberately spans rule-articulable (topical) to
compositionally distributed (sarcasm, deceptive-review style)
signals.

\paragraph{Baselines.}
\label{sec:baselines}
We organize baselines by the cells of Table~\ref{tab:2x2}.
\emph{Anonymous discriminative:}
\textbf{PCA} on the embedding (matched to LFD's $K$ per task);
\textbf{Sparse Probe}~\citep{dalvi2020analyzing}: $\ell_1$-logistic
on the full $3{,}072$-d embedding; \textbf{Full Embedding
($3{,}072$-d)}: LightGBM on the full embedding; \textbf{Fine-tuned
BERT}~\citep{devlin2019bert}: non-linear upper bound.
\emph{Interpretable, non-discriminative:}
\textbf{LDA}~\citep{blei2003latent}, classical probabilistic topic
model, and \textbf{BERTopic}~\citep{grootendorst2022bertopic}, neural
topic model; both forced to $K = K_{\mathrm{LFD}}$ topics per task.
\emph{Interpretable, discriminative (LFD's cell):}
\textbf{TF-IDF + $\ell_1$-logistic} (top-$K$ chi-squared tokens);
\textbf{TBM}~\citep{ludan2024interpretable}, iteratively-generated
single-LLM concept-bottleneck with a linear head; and \textbf{LFD
(ours)}.

\paragraph{LLM stack.} LFD uses three LLM roles, all called at
\texttt{temperature} $= 0$: \emph{proposer} = Grok-4-fast-reasoning;
 \emph{labeler} (applies
admitted features to the training set) =
\texttt{gpt-4o-mini};\emph{examiner} = DeepSeek-chat (independent vendor from the
proposer to avoid same-family blind spots). TBM uses \texttt{gpt-4o-mini} as both
generator and measurer per the published reference implementation.
Cross-vendor $\kappa$ on TBM's admitted concepts
(\texttt{gpt-4o-mini} labeler vs.\ DeepSeek examiner --- the same
probe we apply to LFD features) is reported in
\S\ref{sec:clarity}. Embeddings: \texttt{text-embedding-3-large} ($3{,}072$-d).



\subsection{RQ1: Does LFD Preserve Predictive Utility?}
\label{sec:main_results}

Table~\ref{tab:main} reports \textbf{balanced accuracy}
 (BA) across the ten-task
benchmark.  The anonymous discriminative baselines
 remain stronger raw
predictors on average. This is expected because they do not need to
expose named, independently applicable coordinates. Their role is to
show the predictive signal available when auditability is not imposed
at the coordinate level. The interpretable non-discriminative
baselines, LDA and BERTopic, are weaker on average because their
coordinates are named but not constructed to explain outcome
variation. Within the interpretable group, LFD and TBM are effectively
tied, with a $5$--$5$ split in task-level wins. This is
the intended accuracy result: LFD is not trading away TBM-level
predictive performance to obtain clearer features. Rather, they differ 
in how they reach that peformance.

\begin{table}[t]
\centering
\vspace{-1mm}
\caption{Held-out balanced accuracy across 10 tasks. $K =
K_{\mathrm{LFD}}$ per task; LDA, BERTopic, TF-IDF, PCA are all
forced to this $K$ for matched-rank comparison. We report TBM's BA at
its own admitted $K$ with $K$ in parentheses (e.g.\ ``$0.540$~(9)''
means TBM achieved $\mathrm{BA} = 0.540$ at
$K_{\mathrm{TBM}} = 9$ on AdParaphrase). For each row, cells with
a \colorbox{black!10}{light-gray background} mark baselines that
are weaker than LFD.
Reported BA values are point estimates from a single split per
task; classifier retraining on the fixed admitted basis varies by
$<0.005$ BA across five seeds. This
variance does not include rerunning the LFD proposal pipeline,
which is a more expensive sensitivity question we do not
estimate at this scale. $^{\dagger}$ marks TBM cells whose admitted
basis contains a near-paraphrase concept (max
$|\rho(f, y)| > 0.60$); on these tasks TBM's BA largely
reflects the labeler LLM's zero-shot accuracy on a label-proximal
concept rather than a genuine feature decomposition. \emph{Italicized} TBM cells are both
best in the LFD-vs.-TBM head-to-head and $\dagger$-flagged.\label{tab:main}}
\small
\setlength{\tabcolsep}{3pt}
\begin{tabular}{@{}lrrrcrrrcrrr@{}}
\toprule
 & & \multicolumn{2}{c}{\textit{Int.\ non-disc.}} & & \multicolumn{3}{c}{\textit{Interpretable disc.}} & & \multicolumn{3}{c}{\textit{Anonymous disc.}} \\
\cmidrule(lr){3-4} \cmidrule(lr){6-8} \cmidrule(lr){10-12}
Dataset & $K$ & LDA & BERTopic & & TF-IDF & TBM & \textbf{LFD} & & PCA & Sp.\ Probe & Full Emb. \\
\midrule
AdParaphrase   & 22 & \cellcolor{black!10}0.510 & \cellcolor{black!10}0.515 & & \cellcolor{black!10}0.485 & \cellcolor{black!10}0.540 (9) & 0.665 & & \cellcolor{black!10}0.535 & \cellcolor{black!10}0.585 & \cellcolor{black!10}0.630 \\
Loan Application      &  3 & \cellcolor{black!10}0.745 & \cellcolor{black!10}0.615 & & \cellcolor{black!10}0.695 & \cellcolor{black!10}0.700 (6) & 0.775 & & \cellcolor{black!10}0.670 & \cellcolor{black!10}0.715 & 0.790 \\
AG News S/T    & 26 & \cellcolor{black!10}0.515 & \cellcolor{black!10}0.775 & & \cellcolor{black!10}0.705 & \cellcolor{black!10}0.835 (9)$^{\dagger}$ & 0.855 & & 0.885 & 0.915 & 0.895 \\
Gaslighting    & 17 & \cellcolor{black!10}0.555 & \cellcolor{black!10}0.550 & & \cellcolor{black!10}0.550 & \cellcolor{black!10}0.530 (6) & 0.605 & & 0.610 & 0.620 & 0.630 \\
AG News Spt.   & 33 & \cellcolor{black!10}0.630 & \cellcolor{black!10}0.865 & & \cellcolor{black!10}0.820 & \textit{0.980 (2)}$^{\dagger}$ & 0.885 & & 0.985 & 0.990 & 0.980 \\
AG News World  & 23 & \cellcolor{black!10}0.585 & \cellcolor{black!10}0.770 & & \cellcolor{black!10}0.740 & \textit{0.830 (4)}$^{\dagger}$ & 0.810 & & 0.905 & 0.915 & 0.905 \\
AG News Bus.   & 20 & \cellcolor{black!10}0.615 & \cellcolor{black!10}0.745 & & \cellcolor{black!10}0.680 & \textit{0.795 (6)}$^{\dagger}$ & 0.750 & & 0.855 & 0.895 & 0.855 \\
CFPB           & 21 & 0.595 & \cellcolor{black!10}0.560 & & 0.595 & \cellcolor{black!10}0.550 (5) & 0.565 & & 0.590 & 0.630 & 0.600 \\
iSarcasm       & 15 & \cellcolor{black!10}0.520 & \cellcolor{black!10}0.545 & & \cellcolor{black!10}0.570 & 0.705 (6) & 0.575 & & 0.635 & 0.750 & 0.680 \\
Dec.\ Reviews  & 22 & \cellcolor{black!10}0.530 & \cellcolor{black!10}0.550 & & 0.740 & 0.605 (9) & 0.600 & & 0.855 & 0.870 & 0.860 \\
\midrule
\textit{Mean (10 tasks)} & & \textit{0.580} & \textit{0.658} & & \textit{0.658} & \textit{0.707} & \textbf{\textit{0.708}} & & \textit{0.763} & \textit{0.790} & \textit{0.782} \\
\bottomrule
\end{tabular}
\end{table}

\paragraph{Inspecting LFD and TBM.}
 On several TBM-win tasks, much of
the predictive signal is concentrated in a small number of broad
LLM-judgment concepts: on AG~News~Sports, the single concept
\textit{``mention of athlete or sports team''} alone scores BA
$0.972$ (versus TBM's joint $0.980$); analogous near-saturation
single-concept patterns hold on iSarcasm
(\textit{``mock sincerity''}, $0.713$ vs.\ joint $0.705$) and
AG~News~World (\textit{``Global political and social issues''},
$0.790$ vs.\ joint $0.830$). These concepts are predictive but
illustrate why accuracy alone cannot evaluate named bases: a
concept can work as a classifier input while being close to the
label question. We $\dagger$-flag TBM cells in
Table~\ref{tab:main} where the admitted basis contains a
near-paraphrase concept (max $|\rho(f, y)| > 0.60$ on the test
set) --- AG-News S/T, Sports, World, and Business --- precisely
the rows where the predictive win is mostly the labeler LLM's
zero-shot accuracy on a label-proximal concept. The
basis-size asymmetry (median $K_{\mathrm{LFD}}=21.5$ vs.\
$K_{\mathrm{TBM}}=6$) is a deliberate consequence of LFD's
residual-gain selector: each admitted feature must explain held-out
residual signal without paraphrasing the label, so the basis grows
by narrow sub-cues rather than collapsing onto a few broad ones.
The next two sections test whether the larger basis pays off in
clarity and disentanglement. In the cross-LLM clarity audit,
LFD's features have mean $\kappa=0.97$
($93\%$ clear at $\kappa\geq 0.70$) versus TBM's at mean
$\kappa\approx 0.20$ ($0\%$ clear); applying LFD's $\kappa$-screen
to TBM concepts post hoc retains none. Overall, LFD is most
favorable on tasks that are rule-articulable but not reducible to
a single topic; on compositional tasks (sarcasm, deceptive review)
where the signal lives in context, tone, and interaction effects,
a small named basis is intentionally lossy and anonymous baselines
retain an advantage.

\subsection{RQ2: Are LFD features less label-entangled?}
\label{sec:disentanglement}
We next assess whether named features carry information distinct
from the target label itself. A feature whose definition simply
restates the prediction target may be highly predictive (e.g., ``\emph{mention of sports terminologies}'' for classifying sports news), but it collapses the bottleneck into a renamed classifier and provides
little explanatory value. Existing concept-bottleneck pipelines do
not enforce a disentanglement screen, so their concept sets can
contain a mixture of contentful explanatory cues and concepts that
overlap strongly with the label question. We therefore use three
complementary diagnostics: definition-level similarity to the label
question, labeling-vector similarity to the ground-truth label, and
a human rubric for perceived label leakage. This sharpens the
concept-leakage critique of
\citet{mahinpei2021promises, margeloiu2021concept}: where prior
work flagged concepts that leak \emph{extra-label} information beyond their definition, we identify the dual failure where concepts leak \emph{the label itself}.

\paragraph{Does our method design naturally produce disentangled features?} For the experiments in this paper we set $\tau = 1$, effectively \emph{disabling} the disentanglement gate. The goal is to test whether LFD's structural design --- matched contrastive minibatches, contrastive abductive reasoning over outcome-opposed pairs, and
residual-gain selection --- is by itself sufficient to produce
features that are decoupled from the label, without relying on a
post-hoc filter.
\textbf{Primary metric: $|\rho(f, y)|$
(Def.~\ref{def:interpretable}).} For each feature/concept we compute
the absolute centered cosine between its labeling vector $f$ and the
ground-truth label vector $y$ --- the same quantity the gate $\tau$
would test against. Pooled max-per-task, TBM has median $0.395$ and
LFD has median $0.304$; LFD's largest realized $|\rho|$ on any
admitted feature is $0.59$ (Loan Application,
\textit{``Standard bio template''}), below the $\tau = 0.60$ threshold the gate would have used. By contrast, TBM admits a near-paraphrase concept (max \begin{wrapfigure}{r}{0.5\textwidth}
\centering 
\includegraphics[width=\linewidth,trim=0 0 11cm 0,clip]{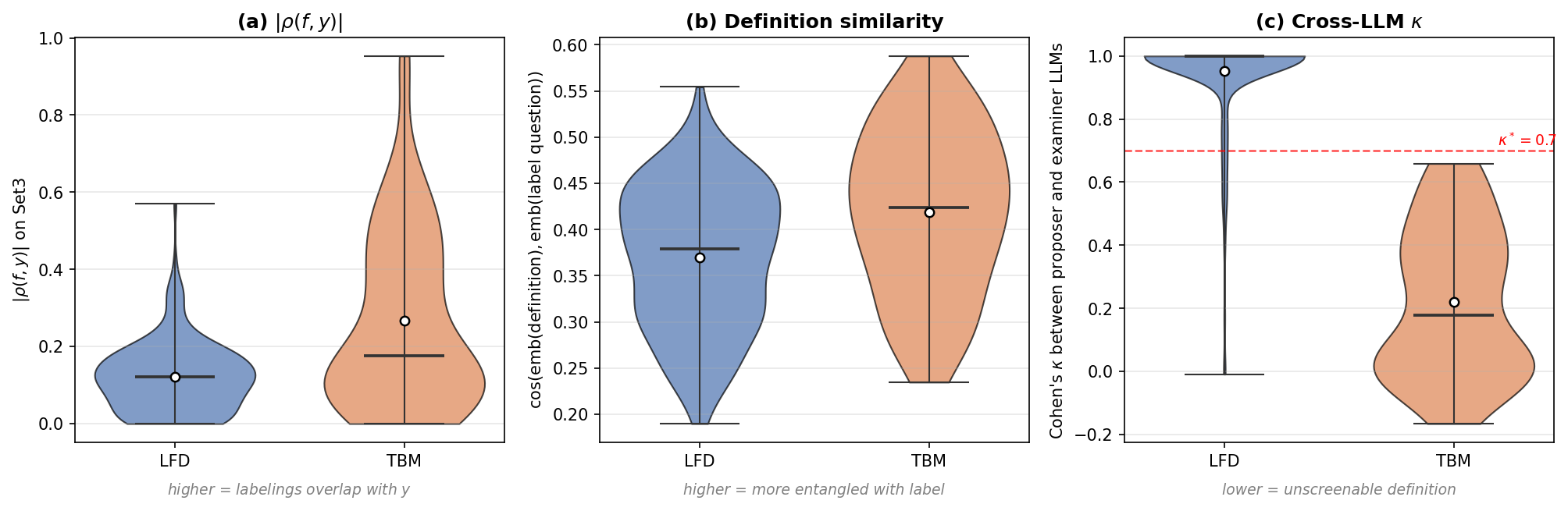}
\caption{Two disentanglement measures of LFD features vs.\ TBM concepts, pooled
across the 10 tasks (\textbf{lower the better}).
\label{fig:disentanglement_boxplot}}
\end{wrapfigure} 
$|\rho(f, y)| > 0.60$ on the test set) on $4$ of the $10$ tasks --- the
$\dagger$-flagged cells in Table~\ref{tab:main} (AG-News Sci/Tech,
Sports, World, and Business). The worst offender is AG-News-Sports'
\textit{``mention of athlete or sports team''} at $|\rho| = 0.94$,
extensionally close to $y$ by construction; on the three
$\dagger$-flagged rows where TBM beats LFD, the reported BA is
largely the labeler LLM's zero-shot accuracy on this label-proximal
concept rather than a genuine feature decomposition.
\textbf{Secondary metric: definition similarity.} We embed each
feature's natural-language description and the task's label question
with \texttt{text-embedding-3-large} and compute cosine similarity
(higher $=$ more entangled at the \emph{definition} level,
complementary to $|\rho|$ which measures entanglement at the
\emph{labeling} level). Both metrics agree:
LFD's design alone produces a materially less label-entangled basis
than TBM's, even without the $\tau$ gate active.

\begin{figure}[h]
\centering

\includegraphics[width=\textwidth]{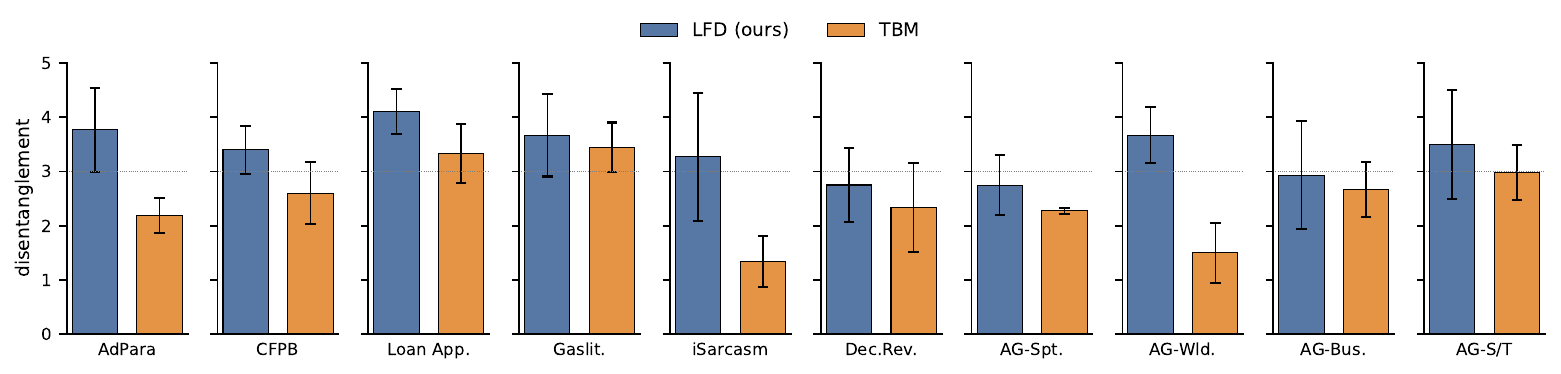}
\caption{Human disentanglement rubric: mean rating per task.
\textbf{Higher is better.} The bar is the across-feature mean of
per-feature means; the error bar is the across-feature SD.\label{fig:rubric}}\vspace{-1mm}\end{figure}

\paragraph{Do human raters find the features label-disentangled?}
We complement the automated metrics with a human rubric: $38$
Prolific raters (English-fluency + $\geq 95\%$
approval) across all 10 tasks (at least $3$ raters per task
who passed an attention check and spent $\geq 5$ minutes) rate each
LFD feature and TBM concept on a 5-point Likert scale from
$1$=\emph{heavy leakage; feature $\approx$ task answer} to $5$=
\emph{no leakage; feature independent of answer}, with two
calibrated reference examples shown in the survey introduction. LFD features score
higher than TBM on \emph{all 10} tasks
(Figure~\ref{fig:rubric}). The largest gaps are AG-World ($+2.17$)
and iSarcasm ($+1.93$), the two tasks where TBM had a predictive
edge over LFD; on iSarcasm, four of TBM's six concepts
(``mock sincerity'', ``subtle irony'', ``exaggerated emotional
tone'', ``emotional inconsistency'') receive the minimum rating.

\subsection{RQ3: Are LFD features conceptually clearer?}
\label{sec:clarity}

We then assess conceptual clarity by measuring agreement among
raters applying the same feature definition. The cross-LLM
$\kappa$ screen used during LFD selection
(\S\ref{sec:proposal}) operationalizes this property with two LLM
annotators, but the operational target is whether independent
\emph{humans} can apply the definition reproducibly. We therefore
measure $\kappa$ at two rater-pair levels: human--human, the target audit; and
human--LLM, a mixed audit of whether pipeline labels align with
human-majority labels.

We collect human ratings on two tasks chosen for opposite
predictive profiles: \textbf{iSarcasm} (TBM beats LFD by $+0.130$
BA) and \textbf{AG-News-Sci/Tech} (LFD beats TBM by $+0.020$). The
contrast tests whether the clarity advantage is a property of the
method or an artifact of where TBM happens to lose. We recruited
$\mathbf{194}$ Prolific raters (English-fluency + $\geq 95\%$
approval) and obtained $\geq 3$ raters per feature after filtering
on attention check + 5-min minimum.

\paragraph{Do human raters find the features conceptually clear?}
For each feature or concept, at least three independent human raters
apply the natural-language definition to the same 25 instances. We
compute Cohen's $\kappa$ between every pair of raters and average
across pairs to obtain the per-feature human--human agreement
 for each method, reported in Table~\ref{tab:human_kappa_summary}. LFD
features are near the substantial-agreement bar of
\citet{landis1977measurement} on both tasks:
$\bar\kappa=0.69$ on iSarcasm and $0.68$ on
AG-News-Sci/Tech. TBM concepts are lower on both tasks:
$0.27$ and $0.45$, respectively. The gap is significant on each
task ($p<0.05$).

\paragraph{Do LLM labels align with human labels?}
The pipeline's labeled feature matrix is produced by an LLM
labeler applying each feature's codebook to every training
instance, so the named coordinates correspond to what humans see
only insofar as human and LLM applications of the same definition
coincide. Human--LLM $\kappa$ measures this directly: LLM labels\begin{wraptable}{r}{0.47\textwidth}\vspace{-2mm}
\centering
\caption{Cohen's $\kappa$ on the human audit.\label{tab:human_kappa_summary}}
\small
\setlength{\tabcolsep}{4pt}
\begin{tabular}{@{}llll@{}}
\toprule
Task & Rater pair & LFD $\bar\kappa$ & TBM $\bar\kappa$ \\
\midrule
\multirow{2}{*}{iSarcasm}
  & human--human  & $0.69${\scriptsize\,$\pm 0.26$} & $0.27${\scriptsize\,$\pm 0.23$} \\
  & human--LLM    & $0.99${\scriptsize\,$\pm 0.03$} & $0.26${\scriptsize\,$\pm 0.12$} \\
\midrule
\multirow{2}{*}{AG-S/T}
  & human--human  & $0.68${\scriptsize\,$\pm 0.28$} & $0.45${\scriptsize\,$\pm 0.19$} \\
  & human--LLM    & $0.84${\scriptsize\,$\pm 0.26$} & $0.43${\scriptsize\,$\pm 0.36$} \\
\bottomrule
\end{tabular}
\end{wraptable} 
align closely with human-majority labels for LFD features
($0.99$ on iSarcasm and $0.84$ on AG-News-Sci/Tech), but much
less so for TBM concepts ($0.26$ and $0.43$). Low human--LLM
agreement does not imply the bottleneck cannot predict the label
--- TBM's classifier does --- but it does mean the named feature
the model trained on differs systematically from the named
feature an independent human auditor would derive by applying the same
description. The codebook becomes a rename of an opaque LLM
judgment, and contestability fails: a human reviewer applying the
codebook to a contested decision will not in general reproduce
the labels the model used. A named coordinate can be predictive
without being conceptually clear; the latter is what auditability
requires.

\paragraph{Empirical anchor for the conditional-independence
assumption.}
Assumption~\ref{ass:annotation_noise}(ii) predicts that if the
two LLMs (labeler and examiner) share residual blind spots,
$\kappa_{\text{ll}}$ should exceed $\kappa_{\text{hh}}$ on the
same features. We test this on AG-News-Sci/Tech, where both
methods admit features whose application is non-trivial. On LFD's
$5$ semantic-lane features with sufficient human ratings,
$\kappa_{\text{ll}}$ exceeds $\kappa_{\text{hh}}$ on $4/5$ (mean
$0.75$ vs.\ $0.39$, gap $+0.36$) --- the shared-bias direction.
On TBM's $9$ concepts the gap is essentially zero ($\approx 0.42$
vs.\ $\approx 0.45$); the iSarcasm cross-check
($\kappa_{\text{ll}} = 0.18 < \kappa_{\text{hh}} = 0.27$) is also
flat. Treating the LFD $+0.36$ gap as a residual co-error rate
$\zeta$ via the \S\ref{sec:theory} relaxation implies an empirical
inflation factor of $\approx 1.30$--$1.45$ rather than the
idealized $1.20$; $\kappa_{\text{ll}}$ is an upper-bound proxy for
$\kappa_{\text{hh}}$ on semantic-lane features. The qualitative
chain (cross-rater $\kappa$ $\to$ bounded per-feature noise $\to$
bounded generalization inflation) holds across this plausible
$\zeta$ range; only the headline $1.20$ constant is optimistic,
and $\kappa^{*} = 0.70$ remains a defensible operating point.


\section{Discussion and Conclusion}

We propose LFD, a feature-discovery procedure designed to satisfy two operational
 interpretability criteria: conceptual clarity (the feature definition is independently
  applicable, measured by cross-rater Cohen's $\kappa$) and label disentanglement (the
  feature does not paraphrase the target). We show that imposing these criteria does not
   substantially reduce predictive utility relative to interpretable competitors.
  And despite similar accuracy, LFD features are substantially clearer to independent human raters and less likely to restate the target label.

  \paragraph{Limitations.} Three caveats bound our claims. (i) LFD
targets rule-articulable tasks; on compositional signals (iSarcasm,
Dec.\ Reviews) the small named basis is intentionally lossy and
anonymous embeddings predict better. (ii) Two transformer LLMs
trained on overlapping corpora may share residual biases that
survive conditioning on the latent feature value; the
$\zeta$-relaxation table bounds the worst
case, and the empirical $\kappa_{\mathrm{ll}}$ vs.\
$\kappa_{\mathrm{hh}}$ audit on AG-News Sci/Tech
(\S\ref{sec:clarity}, Table~\ref{tab:human_kappa_summary})
anchors the gap size on semantic-lane features ($+0.36$, implying
inflation $\approx 1.30$--$1.45$ rather than the idealized $1.20$).
On semantic-lane features $\kappa_{\mathrm{LLM}}$ is an upper-bound
proxy for $\kappa_{\mathrm{human}}$; on simpler features and on
iSarcasm the empirical gap is flat or reversed. (iii) LFD is materially more expensive than its baselines (\$5--15,
3--10 hours per 1{,}000-sample dataset versus $\sim$\$1--2 for TBM
and cents-and-seconds for anonymous baselines); LFD is meant for
settings where auditability is a binding constraint. \paragraph{Broader impacts.} LFD targets high-stakes
text-classification settings (complaint triage, credit
underwriting, content moderation) where decisions must be
inspectable. The same auditability brings two risks: explicit
named bases are easier to evade by adversarial paraphrase than
anonymous embeddings; and passing $\kappa$/disentanglement gates
is not itself a guarantee of fairness or calibration. 


{\small
\bibliographystyle{plainnat}
\bibliography{references,references_may}
}

\end{CJK}
\end{document}